%% file: main.tex
\documentclass{article}
\usepackage[a4paper, total={6in, 8in}]{geometry}
\usepackage[square,sort,comma,numbers]{natbib}
\pdfoutput=1 
\usepackage{graphicx}
\graphicspath{ {./images/} }
\usepackage[dvipsnames]{xcolor} 
\usepackage{url}
\usepackage{comment}

\newcommand{\proj}{\textit{IBM Federated Learning}}
\def\code#1{\texttt{#1}}

\title{
IBM Federated Learning: An Enterprise Framework \\
White Paper V0.1}
\date{\vspace{-5ex}}

 \author{Heiko Ludwig\thanks{Corresponding authors: hludwig@us.ibm.com, baracald@us.ibm.com, gegi@us.ibm.com},  Nathalie Baracaldo, Gegi Thomas, Yi Zhou, Ali Anwar, \\
 	 Shashank Rajamoni, Yuya Ong, 	Jayaram  Radhakrishnan,  Ashish Verma, \\
 	 Mathieu Sinn, Mark Purcell, Ambrish Rawat, Tran Minh, Naoise Holohan, \\
	Supriyo Chakraborty, Shalisha Whitherspoon, Dean Steuer, Laura Wynter,\\
	Hifaz Hassan, Sean Laguna, 	Mikhail Yurochkin, Mayank Agarwal, \\
	  Ebube Chuba, Annie Abay\\ \\ 
	IBM Research}

\begin{document}

\maketitle

\begin{abstract}

    Federated Learning (FL) is an approach to conduct machine learning without centralizing training data in a single place, for reasons of privacy, confidentiality or data volume. However, solving federated machine learning problems raises issues above and beyond those of centralized machine learning. These issues include setting up communication infrastructure between parties, coordinating the learning process, integrating party results, understanding the characteristics of the training data sets of different participating parties, handling data heterogeneity, and operating with the absence of a verification data set.

     {\proj} provides infrastructure and coordination for federated learning. Data scientists can design and run federated learning jobs based on existing, centralized machine learning models and can provide high-level instructions on how to run the federation. The framework applies to both Deep Neural Networks as well as ``traditional'' approaches for the most common machine learning libraries. {\proj} enables data scientists to expand their scope from centralized to federated machine learning, minimizing the learning curve at the outset while also providing the flexibility to deploy to different compute environments and design custom fusion algorithms.
    
\end{abstract}

\newcommand{\alicomment}[1]{\noindent\textcolor{magenta}{\bf Ali: #1}}

\section{Introduction}

\input{intro.tex}

%\section{Understanding Federated Learning Requirements}
%\input{background.tex}
%\section{Context}
%\subsection{Applications}
%\subsection{Research challenges}
%\subsection{Aspects to consider}
%\section{FFL}
%\input{arch.tex}

%\section{Example Use Cases}
%\section{Discussion}

%\section{Related work}
%\subsection{Academic work}
%\subsection{Existing solutions, e.g. FATE, PySyft, TFF}

\input{design.tex}

\section{Supporting different learning paradigms} \label{sec:support-learning-paradigms}

\input{learning_paradigm}

\section{Deployment and Configuration} \label{sec:workflow}
\input{fl_flow.tex}

\section{Conclusion} \label{sec:conclusion}
\input{conclusion.tex}

\section*{Acknowledgements}
It takes the contributions of many to get a system ready for enterprise production. Thanks to Marius Danciu, Jim Rhyness, Frank Li, Calin Radu Coca, Rania Khalaf, Sandeep Gopisetty, Dinesh Verma, and Greg Filla for their contribution to IBM Federated Learning.

\bibliographystyle{plain}
\bibliography{references}
\end{document}

%% file: intro.tex
Federated Learning (FL) is an approach to apply machine learning to situations in which data cannot be centralized for a training process. The success of using machine learning to solve a problem depends, to a large extent, on the quality and quantity of available training data. Machine learning approaches typically rely on the central management of training data, even if the training process itself is run on a clustered infrastructure. The process can consider the properties of the total training data set and the availability of representative validation data sets, e.g., for hyperparameter tuning.
However, centralizing data for training is often not feasible or practical, for reasons of data privacy, secrecy, regulatory compliance, and the sheer volume of data involved.

Enterprises  have their applications deployed in a variety of data centers and clouds. In such a multi-cloud scenario, moving data into one location can be impractical or expensive, particularly if the data are generated at a high velocity or the amount of data is large. A similar situation holds in edge computing scenarios in which, say, data in telecommunication switches is used for training.

% Machine learning models require quality training data in order to achieve high performance in terms of accuracy and generalization. Training data is not always easy to acquire, when training data contains sensitive or confidential information that cannot be revealed to third parties due to privacy, confidentiality or regulatory constraints. 

In recent years, privacy regulations limit the use of personal data, make central data repositories expensive to manage, and represent a liability to the company storing the data. Regulations such as the European Union's General Data Protection Regulation (GDPR) \cite{regulation2016regulation}, the Health Insurance Portability and Accountability Act of 1996 (HIPAA)~\cite{act1996health}, and others require data holders to keep information private and limit how data can be used. Private and confidential data have been compromised from different systems in the past years, with notable security incidents including \cite{facebook:2018:breach,yahoo:breach:2018,google:breach:2018,macy:breach:2018}. These data breaches result in costly mitigation, fines, and reputation damage \cite{yahoo:breach:2018}. 

%The less data is transferred to a central location, the lower the risk of the enterprise learning from data off of mobile applications, medical devices or home automation.

Many privacy regulations also contain limitations on the location where data can be stored. Often, they stipulate that data cannot be taken out of the country to which the regulation applies or at least not to countries that have weaker privacy protections. This even applies to data of the same company if it has, say, customer databases in different countries, and presents a serious obstacle to effective use of data in a training process.

Lastly, in some scenarios, different enterprises cooperate in training a machine learning model, mostly for non-competitive reasons. Financial institutions collaborate to combat money laundering and fraud - as required by regulation - and medical research facilities work jointly on improving diagnostics and treatment design~\cite{suzumura2019towards}. However, due to regulation and secrecy, financial firms will not share transaction data and medical institutions cannot share patient data. 

% Additionally, some law enforcement entities have tried to force companies into providing client's data as it was the case with the FBI-Apple encryption dispute \cite{apple:dispute:2016}. Furthermore, legal precedents show that violating privacy legislation and compromising user’s privacy can result in large monetary fines to compensate for privacy infringements . These fines often trigger big news headlines that negatively impact the reputation of big companies. 

While practical and regulatory reasons limit the centralization of training data, using these different sources of data is still important in many cases to build models on large and representative training datasets. 

% In this climate, privacy requirements are becoming increasingly important while training machine learning models and finding aggregated statistics. One way to overcome the privacy risks is to limit the data shared for training purposes. Consumer companies such as Apple and Google have started new types of solutions to avoid transmitting individual’s private data to a central place while still allowing for the creation of aggregated machine learning models. The advantages of not maintaining users’ private data include avoiding data breaches that could compromise it and their associated liability, not being subject to serve any subpoena for any of its clients avoiding any undesired publicity and limiting the amount of work necessary to fulfill any erasure of data required by legislations such as GDPR.  

 \textbf{Federated learning} (FL) \cite{mcmahan2016google:FL:First} is an approach to train machine learning models that do  not require  sharing  datasets with a central entity. In federated learning, a model is trained collaboratively among multiple parties, which keep their training dataset to themselves but participate in a shared federated learning process. The notion of parties might refer to entities as different as data centers of an enterprise in different countries, compute clusters in different clouds, cell phones, cars, or different companies and organizations.
 
The learning process between parties most commonly uses a single \emph{aggregator}, while peer-to-peer or decentralized models have also been proposed \cite{roy2019braintorrent}. An aggregator would coordinate the overall process, communicate with the parties, and integrate the results of the training process. Most typically, in particular for neural networks, parties would run a local training process on their training data, share the weights of their model with the aggregator, which would then aggregate the weight vectors of all parties using a \textit{fusion algorithm}. Then, the merged model is sent back to all parties for the next \emph{round} of training. 

\textit{Classical machine learning} models can also be adapted to a federated environment. We have to decide which parts of the \textit{federated learning algorithm} would run locally in parties and which would run in an aggregator.
 A federated decision tree algorithm might grow the tree in the aggregator, and query the parties for the count information based on the parties' data sets. Based on these counts, it would compute the information gains for all possible splits and select the feature with the largest information gain to split and built the tree node, and then iterate the next round.
 There are many choices on how to design federated machine learning algorithms, and this is still a very active field of research at the point of writing this paper.
 
FL stands in contrast to established distributed training systems \cite{meng2016mllib,li2014scaling},
where all data is transmitted to a central data center and is subsequently distributed among cluster nodes to train in parallel. This clustered, non-federated approach benefits from understanding the characteristics of the entire training data set and the computational capabilities of the nodes in the cluster, as well as its ability to partition the dataset into convenient chunks among nodes in a cluster. These assumptions do not hold in a federated setting. \\

%\textcolor{blue}{@Heiko I'm not sure what this means: "While the re-engineering of training data from models needs issues to be addressed in federated and centralized cases" 
%My confusion is primarily driven by the fact that the first part of the statement is about training data (I'm not sure what re-engineering means, do you mean pre-processing? Why is this problematic in a centralized case?) and the second part refers to weights.  I modified the phrase below, pls let me know what you think. Thanks!.
%\textbf{Original paragraph:}
%While the re-engineering of training data from models needs issues to be addressed in federated and centralized cases \cite{x}, some federated learning scenarios require that different parties may not derive insights of each others shared model information (e.g., weights) or not even the aggregator may see this information in the plain. Methods of creating differentially private noise or secure multi-party computation are used to meet these privacy requirements. This is mostly the case when data is actually owned by different enterprises or individuals, for example on their phones. Multi-cloud settings might have a lesser need for this. }
%\alicomment{Heiko, does re-engineering cover pre-processing along with techniques used for data privacy in typical distributed ML (non FL)?}

\textbf{Challenges of FL}  arise from different perspectives including data heterogeneity, 
 robustness of the federation process,
 selection of unbiased fusion operators, security and privacy inference prevention and operational and effective deployment in enterprise and multi-cloud settings, among others. 
 While FL does not require the centralized management of training data, thus enabling machine learning to scenarios in which that is not possible, it also poses some new challenges, such as \textit{data set heterogeneity}. In this context, there is no common view of the stochastic properties of the overall training data set, which precludes machine learning dependent on them, established pre-processing methods, fine-tuning algorithms and evaluation of model performance. This includes differences in distribution of data among participants, different dataset sizes for each party, different attributes and general issues of heterogeneity~\cite{chai2019towards}.  \cite{kairouz2019advances} provides a good overview of some challenges of federated learning. 
 
Privacy demands might also add additional restrictions to a federated learning system. For example, while membership inference attacks \cite{inferencefl:melis2019exploiting}, where an adversary may try to infer the training data used during the training process by querying the final model, are possible in both centralized and federated learning settings, additional measures need to be considered in federated learning cases.
Some federated learning scenarios require that different parties are not able to derive insights about each other's training data based on messages exchanged during the training process (e.g., weights).
Additionally, in some other cases, not even the aggregator may be trusted to see this information. Methods of creating differentially private noise or secure multi-party computation are used to meet these privacy requirements \cite{hybrid-alpha,hybrid-one,paillier:damgard2001generalisation,localpd:blum2005practical}. This is mostly the case when data is owned by different enterprises or individuals, an example being data on phones. Multi-cloud settings may have a lesser need for this.
%A framework to enable the seamless selection of different type of security and privacy mechanisms would greatly benefit the enterprise collaborations without requiring enterprise users to be experts in cryptographic techniques.

Another set of challenges relates to needs of the enterprise. (1) FL requires different \textit{skills} from machine learning to distributed systems to cryptography. It's rare to find employees who have all these skills sets. An FL platform must provide a seamless on-ramp for its primary user base, the machine learning professional. (2) An FL platform must deal with \textit{operational complexity}. An FL process requires a more complex infrastructure and processes to enroll parties, distribute models, set up an aggregator, deploy a federated learning algorithm, conduct the federated learning process, and then make a final model available. The system must be resilient to parties with different characteristics and parties dropping out of the training process. (3) \textit{Security and networking} are important to a federated learning solution. Given that FL is typically used where domain boundaries have to be dealt with, networking must suit the deployment needs and facilitate the required security. However, the setup of secure communication should not be burdensome to deploy, avoiding time-consuming processes such as opening ports as much as possible.
 
A few approaches and libraries for FL address some of these issues, including \cite{bonawitz2019towards}, \cite{ryffel2018generic}, and \cite{FATE2020}.
However, {\proj} focuses on enterprise settings where
secure deployment, failure tolerance,
and fast model specification are paramount; these must use existing machine learning libraries that allow enterprise users to take advantage of
a comprehensive set of state-of-the art algorithms without needing to learn new languages. \\

We propose the \textbf{IBM Federated Learning framework} to address these challenges and facilitate an easy integration of FL into productive machine learning workflows in the enterprise. Data scientists and machine learning professionals can benefit from a seamless transition from existing practices of model development to writing federated machine learning models. FL infrastructure must be easy to deploy by an enterprise and fit into the typical operations patterns of IT infrastructure. Finally, researchers in federated learning, those interested in designing novel federated learning algorithms and protocols, can try out their ideas with ease and build on the existing functionality for the specific needs of their organization or the particular application domain. The goal of IBM Federated Learning is to address all of these needs in an easy-to-use framework.

% Inspired by these requirements, we created {\proj} a python-based framework for federated learning in an enterprise environment. It contains a large variety of fusion algorithms and enables a customizable deployment of enterprise-ready communication framework to tailor the framework to the particular requirements of each federation. FFL provides a basic fabric for FL to which advanced features can be added including new fusion algorithms and privacy enhancement techniques. It is agnostic to the specific machine learning library and can be extended to supports different learning topologies, e.g., a shared aggregator.

The remainder of this white paper is organized as follows. In section \ref{sec:concepts}, we present the concepts and terminology.
Then in section \ref{sec:architecture}, we present the architecture of system. 
Section \ref{sec:support-learning-paradigms} demonstrates how to train different types of models;
in particular, we show how neural networks and decision trees can be incorporated into the framework.
In section \ref{sec:workflow}, we show how {\proj} is implemented and how it can be configured, and we conclude in section \ref{sec:conclusion}.

\section{Concepts and Terminology} \label{sec:concepts}
Like any machine learning task, FL trains a model $\mathcal{M}$ over data $D$. $\mathcal{M}$ can be a neural network or any non-neural model. In contrast to centralized machine learning, $D$ is split over $n$ parties, where each party $P_i$ has its own private training dataset $D_i$. An FL process involves 
an \textit{aggregator} $A$ and those $n$ parties $P_1, P_2,..., P_n$ in a way that no party has knowledge of any other dataset than its own. $A$ has no knowledge of any dataset. 

\begin{figure}
    \centering
    \includegraphics[width=0.9\textwidth]{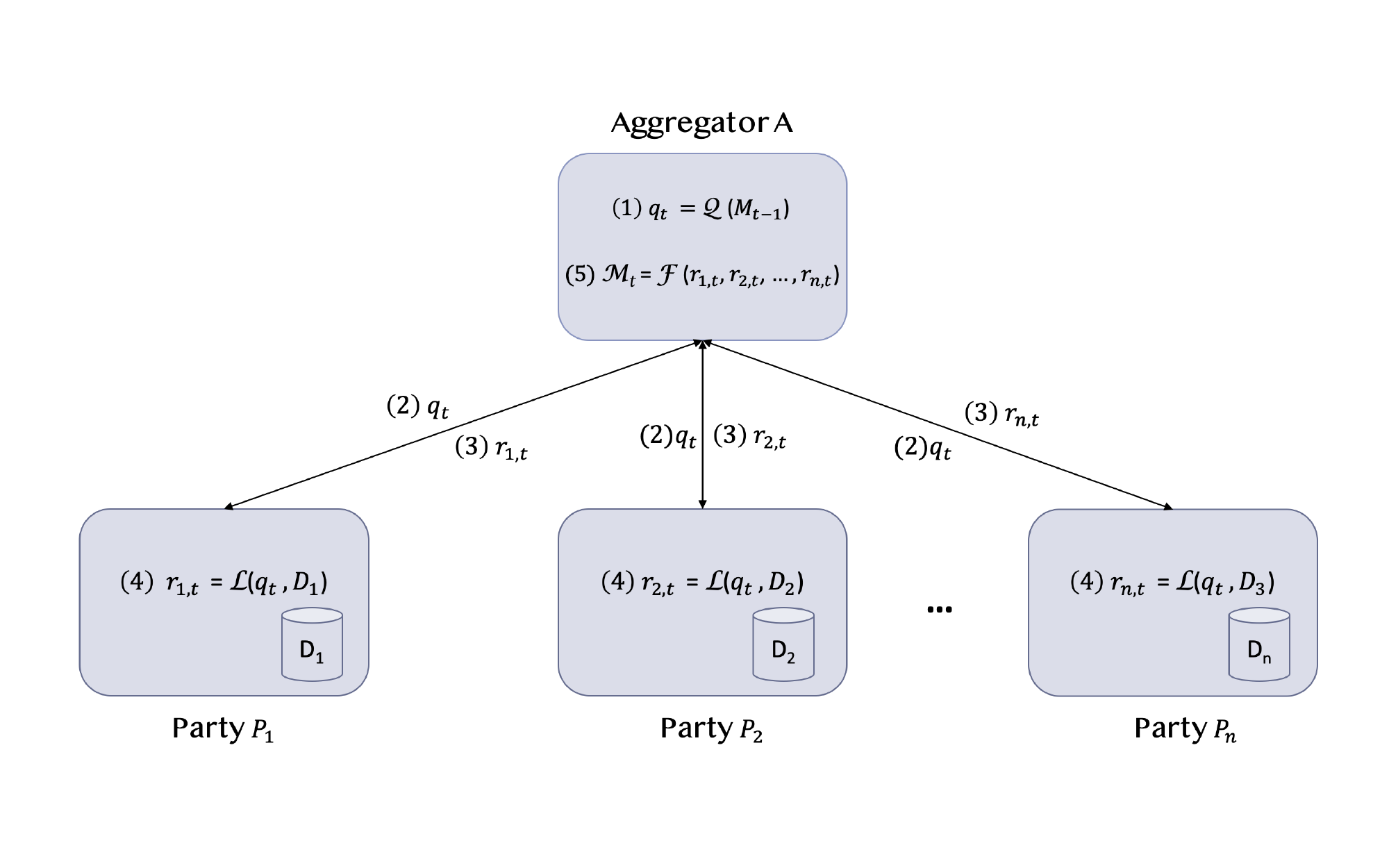} 
    \caption{Federated learning concepts where each party $P_i$ owns its own dataset $D_i$.}
    \label{fig:fl-process}
\end{figure}

The FL process is shown in Figure~\ref{fig:fl-process}.
To train a global machine learning model
$\mathcal{M}_\mathcal{G}$ the aggregator and the parties participate in  a \textit{federated learning algorithm} that is executed by the aggregator and the parties by sending messages. The overall process runs as follows:
\begin{enumerate}
    \item To train $\mathcal{M}_\mathcal{G}$, the aggregator uses a function $\mathcal{Q}$ that takes as input the current model or state of the training $\mathcal{M}_t$ at round $t$,
and generates a next query $q_{t+1}$\footnote{Some FL algorithms may include additional inputs for $\mathcal{Q}$ and may tailor queries to each party, but for simplicity of discussion and without loss of generality, we use this simpler notation.}.

    \item  One such query, $q_t$, requests information about a local model or aggregated information about each party's data set.
Example queries include requests for gradients or model weights of a neural network, or counts for decision trees.
    \item The local training process applies a function $\mathcal{L}$ that takes query $q_t$ and the local dataset $D_i$ and outputs a \textit{model update} $r_{i,t}$. Usually the query, $q_t$, contains information that the party can use to initialize the local training process, for example, model weights to start with local training, or candidate feature values and/or class labels to compute counts for. 

    \item $r_{i,t}$ is sent back from party $P_i$ to the aggregator $A$, which collects all the $r_{i,t}$ from parties $P_i$.
    \item When parties model updates $r_{1,t}, r_{2,t}, ..., r_{n,t}$, 
    where $r_{i,t}$ refers to the model update of party $i$ at round $t$, are received by the aggregator forming set 
    $R_t=\{r_{1,t}, r_{2,t}, ..., r_{n,t}\}$,
    they are aggregated by applying fusion function $\mathcal{F}$ that takes as input $R_t$ and returns $\mathcal{M}_t$.

\end{enumerate}

This process can be executed over multiple rounds $t$ and continues until a termination criterion is met, e.g., the maximum number of training rounds $k$ has elapsed, resulting in a final global model $\mathcal{M}_\mathcal{G} = \mathcal{M}_k$. The number of rounds required can vary highly, from a single model merge of a Naive Bayes model to many rounds of training for typical gradient-based machine learning algorithms. 

The local training function $\mathcal{L}$, the fusion function $\mathcal{F}$, and the query generation function $\mathcal{Q}$ are typically a complimentary set that are designed to work together. $\mathcal{L}$ interacts with the actual dataset and performs the local training, generating the model update $r_{i,t}$. The content of $R_t$ is the input to $\mathcal{F}$ and, thus, must be interpreted by $\mathcal{F}$, which creates the next model, $\mathcal{M}_t$, from this input. If another round $r$ is required, $\mathcal{Q}$ then creates another query. In case of a neural network, for example, $\mathcal{L}$ is the local neural network training algorithm that might produce a weight vector as model update. $\mathcal{F}$ is a fusion algorithm such as a federated average that simply averages each weights over the parties $P_i$, resulting in a new model $\mathcal{M}_t$. For the next round, $\mathcal{Q}$ would then pass $\mathcal{M}_t$ as part of the new query $q_{t+1}$. In this case, $\mathcal{L}$ is computationally expensive and $\mathcal{F}$ less so. If the decision tree is trained in a federated way, one approach might be to balance the decision tree in the aggregator: implement $\mathcal{F}$, send leaf node definitions as part of $q_t$, $\mathcal{L}$ create a response based on a number of data samples corresponding to each leaf node in $D_i$, and then balance the tree again in the aggregator. In this case, $\mathcal{F}$ is computationally expensive and $\mathcal{L}$ is not.

We can introduce different variants to this basic model. While an approach with a single aggregator is the most common and practical for most scenarios, we might choose other configurations. For example, each party $P_i$ might have its own, associated aggregator $A_i$, querying the other parties.
Function $\mathcal{Q}$ might determine the parties to query in each round perhaps based on the merits of prior contributions.
Queries to each party might be different, with $\mathcal{F}$ needing to integrate the results of different queries in the creation of a new model $\mathcal{M}_t$. However, for the further discussion of \proj\ we will focus on FL with a single aggregator.

%% file: design.tex
\section{Architecture}\label{sec:architecture}

{\proj} is a Python library designed to support the machine learning process shown in Figure \ref{fig:fl-process}, while enabling an easy set up in a real distributed environment.

\begin{figure}[hbt]
    \centering
    \includegraphics[width=0.99\textwidth]{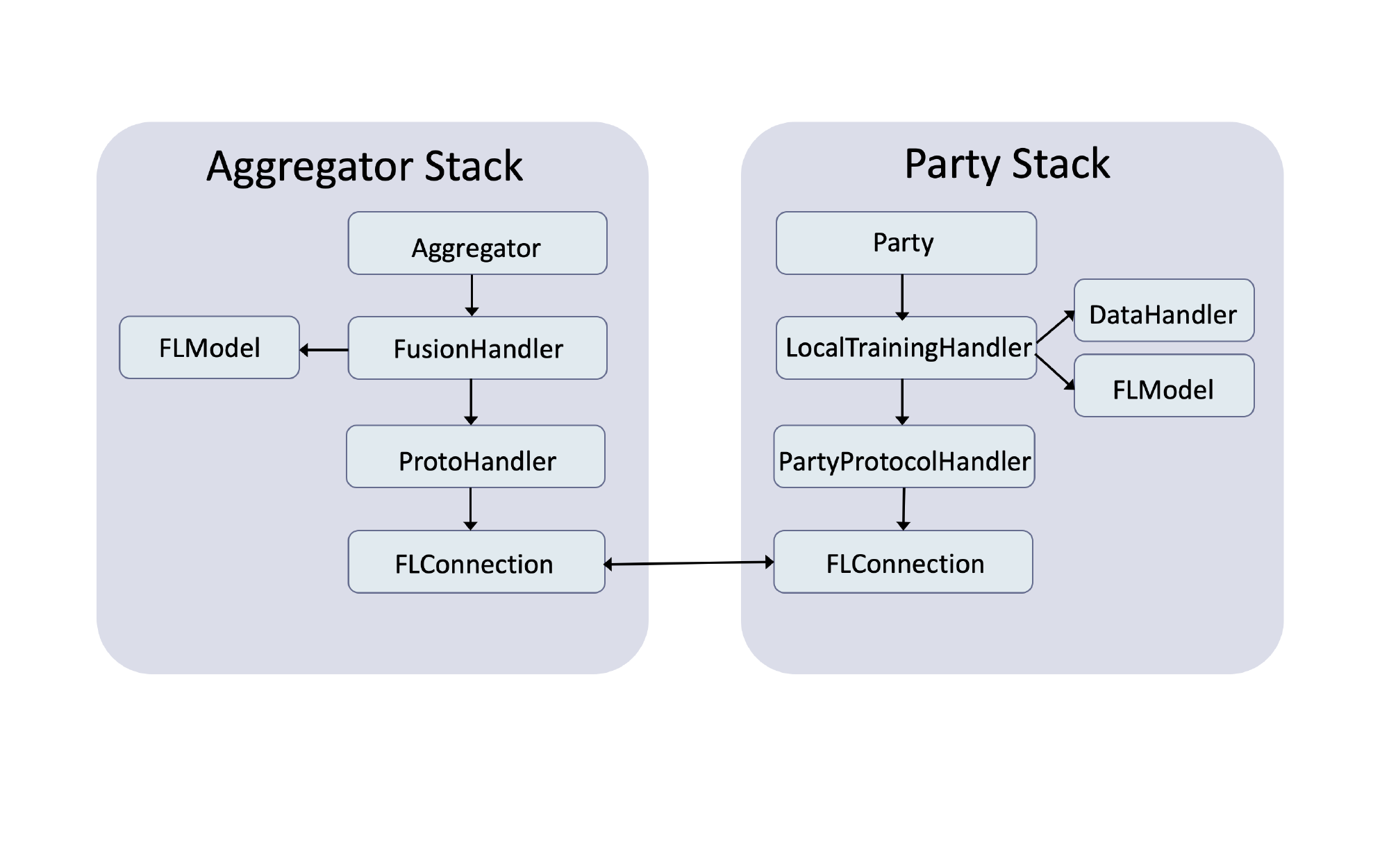} 
    \caption{IBM federated learning architecture stack.}
    \label{fig:ffl-stack}
\end{figure}

{\proj} is designed to implement a resilient platform as well as to ensure  the easy implementation of new FL algorithms.
The {\proj} library contains the components implementing an aggregator $A$ and a party $P_i$ as shown in Figure \ref{fig:ffl-stack}.

This modular design allows the framework to provide communication infrastructure independently from the federated learning algorithm and the actual machine learning library that performs local training. It also enables all parties $P_i$ to read and preprocess data from different locations in different formats.

%he  aggregator  is  in  charge  of  running  the  Fusion  Algorithm.   A  fusion  algorithmqueries  registered  parties  to  carry  out  the  federated  learning  process.   The  queries  sent  varies  onthe model/algorithm types.  In return, parties send back their reply as amodel update object, thesemodel  updates  are  then  aggregated  according  to  the  specifiedFusion  Algorithm,  specified  via  aFusion Handler class.PartyEach party holds its own dataset that is kept to itself and that is used to answer queriesreceived from the aggregator.  Because each party may have stored data in different formats,IBMfederated learningoffers an abstraction called Data Handler.  This module allows for customimplementations to retrieve the data from each of the participating parties.  A local training handlersits at each party to control the local training happening at the party side.

The architecture has the following components: 

\begin{itemize}
    \item \textbf{Connection:}
    Networking is often critical to adoption of any distributed system. The ability to deal with specific requirements and constraints such as available ports, bandwidth, connection stability and latency is important to managing the complexity of deploying a FL system. Multi-Cloud scenarios have different requirements from mobile phone, edge, or company collaboration scenarios. All functionality related to network connectivity for communication between the aggregator and each registered party is handled by the \code{FLConnection} component. {\proj} can support multiple connection types, including the Flask web framework~\cite{flask}, gRPC~\cite{grpc}, and WebSockets.
    
    \item \textbf{Protocol handler:} 
    This component governs  message exchange between parties, i.e. the learning protocol. The message set of the protocol includes a query $q$, a model update $r$, and other messages to establish and dismantle the FL configuration, such as party registration. It uses the \code{FLConnection} for communication and is used by components higher in the stack. It implements which types of messages an aggregator or party can receive at a given point in time. The protocol handler is slightly different for the aggregator and the party stacks due to the differences in messages exchanged. The aggregator contains a \code{ProtoHandler} to manage the protocol to the set of parties $P_1, ..., P_n$, and on the party side, the \code{PartyProtoHandler} implements this function. The party and aggregator side protocol handlers can be adapted to the needs of a particular FL scenario by providing a specialized protocol handler pair.
    
    \item \textbf{Data handler:}
    A \code{DataHandler} is responsible for accessing and pre-processing the local dataset $D_i$ at a Party $P_i$. When running an FL process, the data from each party must be in the correct format so that the learning algorithm can make use of it. While in some FL scenarios, such as mobile phone applications, one will find the same data structure at each party, enterprise scenarios often face different data structures in applications in different clouds or in data centers of different companies. The \code{DataHandler} provides the abstraction to import data specifically for a party. The \code{DataHandler} is designed as an easily customizable module for each party to prepare their data to run. Other {\proj} modules will only interact with the data handler via limited APIs, for instance by \code{{\sl get\_data}} to access training and testing data sets.
    
    \item \textbf{FL training modules:} 
    FL machine learning algorithms are incorporated into {\proj} by specifying a \code{FusionHandler} and a \code{LocalTrainingHandler} for the aggregator and parties, respectively.
    
    \begin{itemize}

        \item \code{FusionHandler}: This module contains the specification of functions $\mathcal{Q}$ to generate queries and $\mathcal{F}$ to fusion model updates. The module uses high level APIs provided by the \code{ProtocolHandler} to send and receive messages generated by $\mathcal{Q}$ and also obtain all party replies to aggregate model updates using the specified $\mathcal{F}$ function.
        Different implementations of $\mathcal{Q}$ and $\mathcal{F}$ may require different information to be exchanged. For example, in some $\mathcal{Q}$ implementations a message may be added in addition to the query hyperparameters. To accommodate this differences of the information exchanged by the \code{FushionHandler}, {\proj} makes use of a dictionary, a generic data structure with customizable fields. 
        
        \item \code{LocalTrainingHandler}: This module is used to specify function $\mathcal{L}$ that will be run at the parties' side, to generate model updates and send them to the aggregator. The \code{LocalTrainingHandler} may use a \code{FLModel}.
        
        \item \code{FLModel}: This module provides an interface to define a standard API to train, save, evaluate, and update the model, as well as generate model updates to be specified in different ML libraries, such as Keras or scikit-learn.
        For each ML library, a module is instantiated to wrap all the offered API (e.g., evaluate, train, save, update model).
        In this way, the \code{FusionHandler} and the \code{LocalTrainingHandler} can use the standard API and, as a result, be used without changes for multiple supported ML libraries.
        %Similarly to DataHandler, other {\proj} modules only interact with FLModel via limited APIs, e.g., {\sl fit\_model}, {\sl update\_model}, and {\sl evaluate\_model}, and so on. 
        %{\proj} has a LocalTrainingHandler module to invoke certain functionality of FLModel, for example, updating and/or training the local model, and then construct a model update based on the current local model and/or local dataset at the party side. Only PartyProtoHandler accesses LocalTrainingHandler.
    \end{itemize}

    %{\proj} has an additional component, namely \textbf{hyperparameters}, for users to specify the hyperparameters for fusion algorithms, which includes two parts: 1) global training hyparameters used for fusion handler and local training hyperparameters utilized by parties to perform local training.
    % \item \textbf{Hyperparameters:} Global training and local training hyperparameters.

\end{itemize}

%\paragraph{Aggregator} The aggregator is in charge of running the Fusion Algorithm. A fusion algorithm queries registered parties to carry out the federated learning process. The queries sent varies on the model/algorithm types.  In return, parties send back their reply as a \textit{model update object}, these model updates are then aggregated according to the specified \textit{Fusion Algorithm}, specified via a \textit{Fusion Handler class}. 

%\paragraph{Party}
%Each party holds its own dataset that is kept to itself and that is used to answer queries received from the aggregator. Because each party may have stored data in different formats, {\proj} offers an abstraction called Data Handler. This module allows for custom implementations to retrieve the data from each of the participating parties. A local training handler sits at each party to control the local training happening at the party side.
%\\

%\subsection{Architecture}
%Figure~\ref{fig:ffl-stack} shows the overall architectures of {\proj} which has been color-coded to show how different components are paired.
%Both aggregator and party sides have their own protocol handlers, connections, models, and data handlers. In addition to that, the aggregator has an additional component called "fusion" whereas party has "local training" module. 

\subsection{Aggregator stack}

The aggregator stack contains the components implementing an aggregator $A$. Its role includes the coordination of the federated learning process, the execution of $\mathcal{F}$, the fusion algorithm, and persisting the  meta-data of the FL process.

The metadata about a federated learning process includes information such as the list of parties, the current state of the process, logging information, and aggregated models $\mathcal{M}_t$. It also performs management tasks such as starting and closing the training process.

As shown in Figure~\ref{fig:ffl-stack}, the aggregator works with the \code{ProtoHandler} and the \code{FLConnection} to communicate with data parties. The \code{FusionHandler} serves to generate queries and to aggregate received replies. In some cases, it is desirable to know the performance of a global model during the training process, for instance to evaluate early termination criteria. In such cases, the aggregator may have access to some testing samples which are accessed through an optional \code{DataHandler}.

\begin{figure}[htb]
	\centering
	\includegraphics[width=0.99\textwidth]{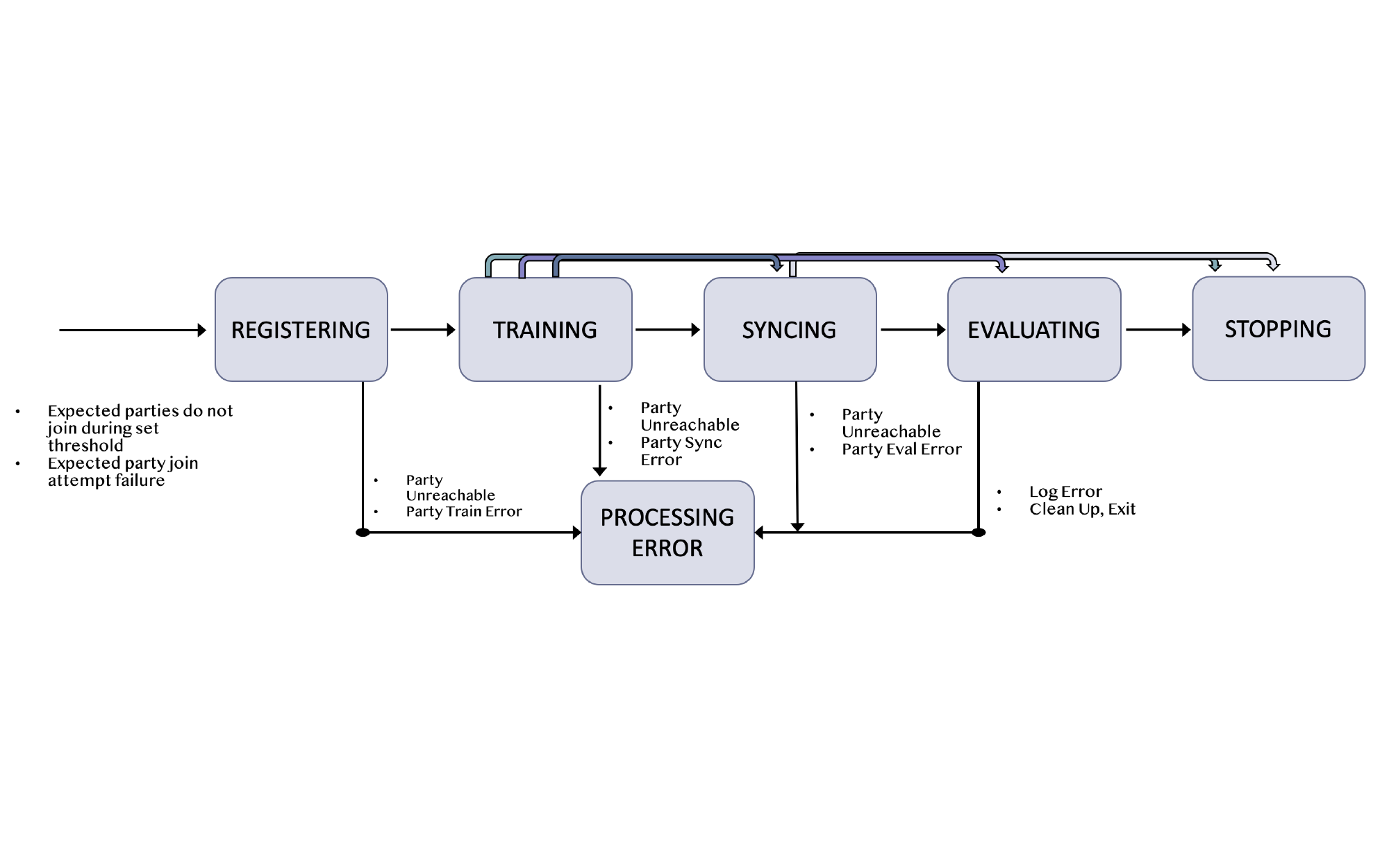} 
	\caption{Aggregator phases. }
	\label{fig:agg-states}
\end{figure}

The aggregator provides an interface to the users to issue commands to control the overall federated learning process.
Figure~\ref{fig:agg-states} shows different phases of the aggregator which include \code{REGISTERING}, \code{TRAINING}, \code{SYNCING}, \code{EVALUATING}, \code{STOPPING} and \code{PROCESSING\_ERROR}.

Once the aggregator is started, it waits for the data parties to register. Once the registration process is completed and there is a quorum, a user can issue a \code{TRAINING} request to start the training process, which triggers the execution of the \code{FusionHandler}. During this process, the aggregator uses the \code{ProtoHandler} and the \code{FLConnection} to send the queries generated by the \code{FusionHandler}. Each party responds to this request by sending the trained model updates. Making use of the \code{FusionHandler}, the aggregator computes $\mathcal{F}$ to aggregate all model updates and generate the queries for parties. This process is repeated until we reach a termination criteria, i.e. desired number of training rounds or specific model accuracy.

Once termination criteria is reached, the aggregator sends a final model update to all the parties via a \code{SYNCING} command which distributes the final global model $\mathcal{M}_\mathcal{G}$. Next, the aggregator may request parties to perform a local evaluation before stopping the federated learning process.
If an error arises during any of these steps, the aggregator logs it and stops the federated learning process. {\proj} collects metadata that allows for failure recovery by going to a previous valid phase.

\subsection{Party stack}
Similar to the aggregator, the party side consists of its own protocol handler, connection, model, local training, and data handler. Unlike the aggregator, the \code{DataHandler} is mandatory for parties. The party provides a command interface for users to start interacting with the application.
Each party starts with issuing a \code{REGISTER} command to join a federated learning process. After the registration process is completed, the party waits for the aggregator to send a message with a query $q$ to start running the \code{LocalTrainingHandler}, which executes $\mathcal{L}$ based on the received $q$ and produces a reply $r$.
Then, $r$ is sent back to the aggregator. This process is repeated until the aggregator sends a request to stop the federated learning process.

In some cases, the aggregator may also issue a \code{SYNC} request, which includes the final global model $\mathbf{M}_\mathcal{G}$.

%% file: learning_paradigm.tex
In the past few years, a variety of algorithms have emerged for training different types of ML models via federated learning, e.g., \cite{mcmahan2016google:FL:First,tifl:chai2020,hybrid-alpha,hybrid-one,pfmn:yurochkin2019bayesian,SPAHM:yurochkin2019statistical,krum:blanchard2017machine,xie2019zeno,goel2020coordinate}. 
Since FL is an approach to perform collaborative learning with the help of an aggregator for coordination, existing FL algorithms usually contain two parts:
1) A global training module, the \code{FusionHandler}, mainly contains the fusion function $\mathcal{F}$, and the query generation function $\mathcal{Q}$;
2) A local training module, the \code{LocalTrainingHandler},  contains the local training function $\mathcal{L}$.  
They come in pairs to execute a FL algorithm in {\proj}.
However, different federated learning algorithms vary significantly in several aspects: how and where
the global model $\mathcal{M}_\mathcal{G}$ is updated;
the query generation function $\mathcal{Q}$;
the fusion function $\mathcal{F}$;
the function $\mathcal{L}$ preformed by the parties and
the content of model update $r$, among others.
{\proj} is designed to be flexible enough to accommodate different learning paradigms in FL. 
We cover the design of several different FL algorithms for {\proj} below.

% content list
% \begin{itemize}
%     \item Supporting neural networks specified in Keras (here explain two fusion algos.)
%     \item Supporting decision trees
%     \item RL 
%     \item Bayes naives 
% \end{itemize}

% \item Supporting neural networks specified in Keras (here explain two fusion algos.)
\subsection{Neural networks}\label{sec:nn}

In this subsection, we focus on the discussion of training neural networks via iteratively updating $\mathcal{M}_\mathcal{G}$ by the (weighted) average of local models' parameters.

In particular, this type of FL algorithm requires the aggregator to generate a query, $q_t$, containing the current global model weights - and hyperparameters, optionally - and send it to all or a subset of parties at round $t$.

This query is implemented as a dictionary.
Once parties receive $q_t$,
each party initializes its local model with the global model weights, performs several epochs of local training with the received hyperparameters, if any, to obtain the model update $r_t$.
The resulting $r_t$ contains the new set of local model weights, and shares it with the aggregator.
In some variations, where a weighted average is used ({\sl FedAvg}~\cite{fedavg:mcmahan2017communication}),
the number of data points used for training at each party is included in $r_t$.
The model update $r_t$ is implemented as a dictionary, and so, when {\sl FedAvg} is executed, two key values are included into the dictionary with key values ${weights, nsamples}$. In some cases, for privacy considerations, the number of samples at each party cannot be transmitted to the aggregator. In those cases, the model update only has a single-item dictionary.
The aggregator then follows the corresponding fusion function $\mathcal{F}$ to conduct a (weighted) average
over the collected $r_{i,t}$'s and updates the global model's weights with the fusion results. 
Both the global and local models' structure, i.e., the neural network architecture, stay the same.

%
%We have discussed in Section~\ref{sec:architecture}, the \code{FusionHandler} performs the global training at the aggregator side.

What does this mean for the \code{FusionHandler}? 
Implementing the aforementioned type of FL algorithms, \code{IterAvgFusionHandler}, a subclass of \code{FusionHandler}, which, after being invoked by the aggregator, iteratively constructs a query containing the current weights of $\mathcal{M}$, and sends the query via the API offered by the protocol handler; this, in turn utilizes the connection layer to send the messages to parties upon receiving enough reply updates $\mathcal{M}_\mathcal{G}$ with the fusion results of $\mathcal{F}$.
Depending on the fusion algorithms, the fusion function $\mathcal{F}$ can be a simple average of the collected $r_{i,t}$'s as implemented in \code{IterAvgFusionHandler}, or a weighted average whose weights depend on the parties' sample size as implemented in \code{FedAvgFusionHandler}.
Note that the latter fusion function expects the replies $r_{i,t}$ to contain the parties' sample size in addition to the local model's weights.

The \code{LocalTrainingHandler} module implements the local training function $\mathcal{L}$ on the party side.
The \code{LocalTrainingHandler} triggers the party to perform a predefined number of training epochs on the local model, and constructs a model update $r$ containing the information requested by the corresponding \code{FusionHandler}. 
For example, a model update replying to the query constructed by \code{IterAvgFusionHandler} will contain the new set of local model weights, while a model update replying to the query constructed by \code{FedAvgFusionHandler} will contain the local sample size along with the local model weights.

% \textcolor{blue}{I added this in detail in the first paragraph mentioning the dictionary... let's revisit to see in which paragraph flows better.}

{\proj} employs the \code{FLModel} module to allow the users to provide model definitions in standard ML libraries and to provide a single API to a pair of a \code{FusionHandler} and a \code{LocalTrainigHandler}.
As a concrete example, a data scientist can specify a convolutional neural network using the standard Keras library \cite{chollet2015keras}.
\code{KerasFLModel} wraps the functionality for
initializing the model according to the specified Keras model definition, 
training, 
creating a model update (extracting model weights from the neural network)
and
saving the final model.

Finally, it is worth mentioning that {\proj} also includes other \code{FLModel} modules. Our design ensures that the same FL algorithm, i.e., the same pair of a \code{FusionHandler} and a \code{LocalTrainigHandler}, can be used to train different models specified using different ML libraries.
For instance, the two FL algorithms we discuss above can also be applied to train Scikit-learn \cite{scikit-learn} linear models in {\proj}.

% \item Supporting decision trees
\subsection{Decision trees}

Training a decision tree in a federated setting requires a different approach to implementing a federated learning algorithm than neural networks do. We illustrate this for an adapted ID3 algorithm.
Since this type of decision tree deals with discrete feature values, its FL algorithm is very different than those we have discussed in the previous section.

{\proj} supports a FL variant of the ID3 algorithm \cite{quinlan1986induction}, where the tree is grown at the aggregator side and the local training function, $\mathcal{L}$ only performs light computations to generate replies $r$ containing the counts information.
No initial model structure is provided to the aggregator and parties, and thus the aggregator starts with a null tree root node.

During the training process, the query generation function $\mathcal{Q}$ takes the current list of candidate feature values for splitting and class labels to query the parties for their counts information. 
The local training function $\mathcal{L}$ computes the corresponding counts based on the party's local dataset, and the resulting model update $r$ is shared with the aggregator.
The fusion function $\mathcal{F}$ splits the current tree node according to the information gain  computed based on the collected counts. 
When the tree reaches the maximum depth or all tree nodes have no candidate feature values to split, the training process ends.

The \code{ID3FusionHandler}, invoked by the aggregator, implements the core parts of the ID3 FL algorithm, i.e., the fusion function $\mathcal{F}$ and query generation function $\mathcal{Q}$.
$\mathcal{Q}$ recursively takes the updated list of candidate feature values, and constructs the query $q$ with the candidate list.
The query $q$ is sent the same way to parties as by another \code{FusionHandler}.

After collecting a quorum of replies $r_{i,t}$, i.e., the list of counts sent by the parties, the fusion function $\mathcal{F}$ computes the sums of the counts over all parties with respect to their corresponding feature values and recovers the overall counts for the candidate feature values.
Following the information gain formula used by the ID3 algorithm, $\mathcal{F}$ chooses the feature value with the maximum information gain to split and construct the resulting tree nodes according to the predefined criterion, whether reaching the maximum tree depth or having a empty candidate list.
It also updates the candidate list and moves on to the new constructed tree node. 
When the training process finalizes all leaf nodes, the tree is grown at the aggregator side.

As we can see \code{ID3FusionHandler} performs the computation-heavy operations at the aggregator, different from the two FL algorithms we have discussed in Section~\ref{sec:nn} to train neural networks where the computation expensive training happens at the party side.
The \code{LocalTrainingHandler} triggers a light operation conducted by the parties to compute counts information and constructs the model update $r$ containing these counts.

In the case of an ID3 decision tree, {\proj} implements the \code{DTFLModel} independently of any existing ML library.
It follows the same convention as other \code{FLModel} to implement a list of methods, such as \code{\sl fit\_model}, \code{\sl update\_model}, \code{\sl get\_model\_update}, \code{\sl evaluate\_model} and \code{\sl save\_model}, etc., so that other modules can interact with it the same way as with other types of ML models.

% \yi{TBA}
% content list
% \begin{itemize}
%     \item Supporting neural networks specified in Keras (here explain two fusion algos.)
%     \item Supporting decision trees
%     \item RL 
%     \item Bayes naives 
% \end{itemize}

\subsection{Other models, differential privacy and multi-party computation}

The flexible architecture of {\proj}  also supports the implementation of other FL algorithms as well as the use of differential privacy and multi-party computation. In the following, we discuss how to use these approaches are implemented based on this library. \\

\noindent \textbf{Other models and training algorithms:}
{\proj}  supports the training of different machine learning models besides neural networks and decision trees. 
These include adaptations of XGBoost for binary and multi-class classification as well as regression,
% \cite{our:fl:xgboost},
linear classifiers and regressions with regularizers including logistic regression, linear SVM, ridge regression and more.
The framework also comes with Naïve Bayes and
Deep Reinforcement Learning algorithms including DQN, DDPG, and  PPO, among others.

With respect to FL algorithms, several common and advanced algorithms are now part of the framework's algorithm library.
The set of common algorithms  includes average  and weighted aggregation, FedAvg \cite{fedavg:mcmahan2017communication}, as described above.
We have implemented multiple advanced algorithms including SPAHM \cite{SPAHM:yurochkin2019statistical} which uses a statistical model aggregation via parameter matching for supervised and unsupervised learning,
%FedMA \cite{Wang} --> not in FFL yet
and PFNM \cite{pfmn:yurochkin2019bayesian}, which provides a communication efficient method for federated learning of fully-connected networks with adaptive global model size.

Algorithms to improve robustness such as Krum \cite{krum:blanchard2017machine}, 
a Byzantine tolerant gradient descent coordinate-wise median fusion \cite{goel2020coordinate},
and
Zeno, a distributed stochastic gradient descent with suspicion-based fault-tolerance \cite{xie2019zeno} are also available.
New algorithms, as needed, can be easily implemented within the existing framework and added back to the library.  \\

\noindent \textbf{Privacy-aware algorithms}
Real federated learning scenarios may require different techniques to ensure the federated learning process can be performed.

In particular, in some cases, participants may have limited trust amongst each other or in the aggregator.
In these cases, it is imperative to add additional protection mechanisms, including secure multi-party computation and differential privacy.

The right technique to apply largely depends on the threat model of choice.
Techniques such as local differential privacy \cite{localpd:blum2005practical},
multi-party computation (SMC) \cite{paillier:damgard2001generalisation,hybrid-alpha} and hybrid approaches have been proposed for this purpose \cite{hybrid-one}.

%\noindent \textit{Secure Multi-party Computation}
Secure multi-party computation techniques allow for the private aggregation of inputs among parties, where a function value can be computed without revealing the function's inputs.
Among relevant algorithms for this purpose include \cite{paillier:damgard2001generalisation}, which utilizes partial homomorphic encryption,
\cite{hybrid-alpha}, which makes use of functional encryption to drastically speed up the training process in comparison to homomorphic-based approaches,
and \cite{hybrid-one}, where differential privacy and SSM (threshold-based Pailler scheme) are combined to create highly accurate models without compromising the offered differential privacy guarantee.

Some of the previously mentioned SMC techniques require a different number of messages exchanged. For instance, while the approach presented in \cite{hybrid-alpha}, based on functional encryption, only requires the aggregator to send a single query, $q$ and participants to send a single encrypted model update, $r$,
techniques like \cite{hybrid-one} require multiple communication rounds between the aggregator and parties.
Ideally, all these nuances should be hidden, ensuring that a crypto-aware
 \code{Fushion\_Handler} and \code{LocalTrainingHandler} do not need to be specifically adapted for each type of cryptosystem

{\proj} provides a simple API to incorporate a diverse set of cryptosystems that enable machine learning professionals to invoke crypto operations in the \code{Fushion\_Handler} and \code{LocalTrainingHandler}.
We designed this part of the system to ensure cryptosystems can be interchanged without modifying a crypto-aware pair of a \code{Fusion\_Handler} and a \code{LocalTrainingHandler}.

We have incorporated several SMC techniques including
partial homomorphic encryption \cite{paillier:damgard2001generalisation},
the approach proposed in \cite{hybrid-alpha}
and the functional approach presented in \cite{hybrid-one}.

{\proj} also provides the building block to add differential privacy for different types of models varying from very simple Naive Bayes to more complex differential privacy mechanisms as those required for neural networks. For the first type of cases, a one-shot differential privacy addition that can be applied by a mechanism at the \code{LocalTrainingHandler}.
For other models, an accounting module is added at the aggregator side to keep tract of the differential privacy budget.
Multiple variations on differential privacy methods may be implemented including \cite{mironov2017renyi:dpaccounting,abadi2016deep}. \\

\noindent \textbf{Machine learning libraries}
Finally, {\proj} was designed to ensure ML professionals can specify their models using common ML libraries.
Our de-coupled design ensures that implemented \code{FusionHandler}s and \code{LocalTrainingHandler}s that use the APIs offered by \code{FLModel} can be reused for different libraries. 
Currently, we include  \code{FLModel} implementations for
Keras,
Pytorch,
Tensorflow,
Scikit-learn and
RLlib.
This list can be extended by users implementing the interface of  \code{FLModel} for other libraries.

%% file: fl_flow.tex
%In this section, we will walk through the process of running a federated learning job using IBM federated learning.

%As aforementioned, {\proj} offers as abstraction an \textit{Aggregator} and \textit{Party} so that a group of parties under the coordination of an aggregator can collaboratively train a machine learning model without revealing their raw data. 
%Only \textit{updates}, e.g., model weights or gradients, are shared with the aggregator. 
%The aggregator then updates and learns a global model by aggregating the information provided by the parties. 
% We now introduce these two abstractions in IBM FL.
% \textcolor{blue}{Configuring FL via YAML file, and start FL training flow}

Deploying the {\proj} environment involves deploying an aggregator stack and a party stack at each of the environments involved. Since {\proj} is a Python library, its deployment follows the same workflow as other Python packages, configuring its dependencies. The party library is typically installed where the local training will take place.

{\proj} is designed to allow different modules to be swapped in and out without interfering with the functionality of other components. This requires users to configure the choices that work for their configuration. Party and aggregator stacks can be configured via an external API configuration, or driven by a configuration file (YAML), which are different for the aggregator and each party. {\proj} provides scripts to generate configuration files with default settings which users can modify according to their requirements.

We now briefly discuss a list of building blocks that can be configured in {\proj}.

\vspace{.1in}
\noindent {\bf Connection}: It is configured to provide information needed to initiate the connection between the aggregator and parties. For example, flask server information, synchronization mode for training requests, Transport Layer Security (TLS) configurations, etc.

\vspace{.1in}
\noindent {\bf Data}: It is configured to provide information needed to locate and load the local data sets and perform any pre-processing techniques.

\vspace{.1in}
\noindent{\bf Federated learning algorithm}: {\proj} supports a variety of FL algorithms to train different types of machine learning models. Configuring a federated learning algorithm usually contains two parts:
1) \code{FusionHandler} is configured to provide information about the Fusion algorithm used at the aggregator;
2) \code{LocalTrainingHandler} is configured to provide information about the $\mathcal{L}$ function.

% \vspace{.1in}
% \noindent{\bf Local training}: It includes information needed to initiate a local training at the party side.

\vspace{.1in}
\noindent{\bf Hyperparameters}: It is configured to set up global and local training hyperparameters, e.g., global training round number, local training epochs, batch-size, learning rate, etc.

\vspace{.1in}
\noindent{\bf Model}: It is configured to provide information needed to initiate a machine learning model, which can be defined via existing ML libraries, like Keras and Scikit-learn among others, or a self-implemented model class.

\begin{figure}[bth]
	\centering
	\includegraphics[width=0.90\textwidth]{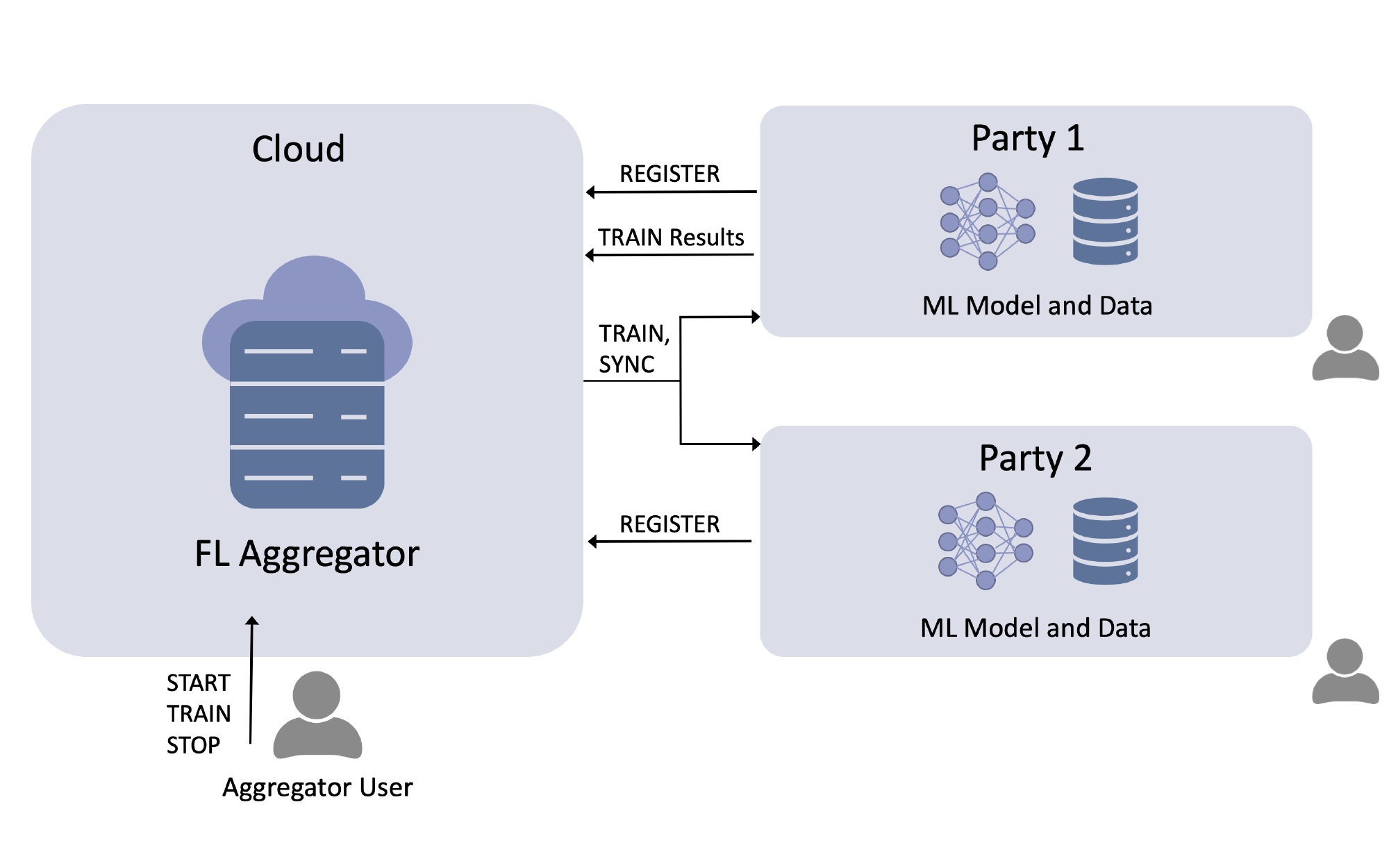}
	\caption{Flow to execute a FL job using {\proj}.}
	\label{fig:ffl-arch}
\end{figure}

\vspace{.1in}
\noindent{\bf Protocol  handler}: It is configured to provide information needed to initiate a protocol which oversees the message exchange between the aggregator and parties. 
\\

After setting up the FL job details in configuration files, users can run a federated learning job in {\proj}.
% \textcolor{red}{Walk through how to set up a federated learning job.}
%
%

The aggregator and party stacks come with a main application to run independently, or can be integrated into another application. To start the aggregator, a user sets up the proper environment with {\proj} installed, and launches the aggregator application with the aggregator configuration file.
As shown in Figure~\ref{fig:ffl-arch}, the aggregator application can be launched in a cloud or cluster environment. 

Similarly, the party applications are launched in an environment with {\proj} installed and the individual parties' configuration files are provided to specify the FL job settings.
After the aggregator application starts running, parties can register themselves via the \code{REGISTER} command. The aggregator waits for a quorum to start the FL training process using the \code{TRAIN} command. 

The following table provides an overview of the commands of  aggregator and party. In the applications included in the library they can be issued on the command line.

\begin{table}[!htbp]
	\centering
	\caption{A list of commands for {\proj}}
	\vspace{.1in}
	\label{tab:fl_commands}
	\begin{tabular}{|c|c|c|}
		\hline
		\textbf{IBM FL Command}      & \textbf{Participant}                                                             & \textbf{Description}                 \\ \hline
		{ \code{START}} & aggregator / party &	Start accepting connections          \\ \hline
		{\code{REGISTER}} & party & Join an FL job
		\\ \hline
		{ \code{TRAIN}} & aggregator                                                                       & Initiate training process            \\ \hline
		{\code{SYNC}}  & aggregator                                                                       & Synchronize model among parties      \\ \hline
		{\code{STOP}}                         & aggregator / party                                                                       & End experiment process               \\ \hline
		{\code{EVAL}}                         & party                                                                            & Evaluate model                       \\ \hline
		{\code{SAVE}}   & {aggregator / party} & Save the model locally at party side \\ \hline
	\end{tabular}
\end{table}

The users can observe the FL training process via log information. 
Once the FL training finishes, the main applications provide a list of commands to be used to collect the FL results, see Table~\ref{tab:fl_commands}. If the library is embedded in an application it can be controlled and monitored in its contained application context.

%% file: conclusion.tex
This white paper introduces {\proj}, a Python framework for federated learning in the enterprise. This framework is a platform for executing federated learning algorithms by providing all of the basic elements of a federated learning infrastructure in an easily configurable way. On the one hand, it provides hooks to implement the party and aggregator parts of a federated learning algorithm. On the other hand, it is configurable to a variety of deployment scenarios, from mobile and edge scenarios, to multi-Cloud environments in an enterprise, to use cases across organizational boundaries. {\proj} is independent of a particular machine learning library or machine learning paradigm. It can be used for deep neural networks as well as ``traditional" approaches such as decision trees or support vector machines. The abstractions for different model types provide a high level of reuse of federated learning algorithms for different machine learning libraries. A fusion algorithm for a deep neural network in Keras works as well with a native PyTorch one. Data access can be defined for each party independently. This is particularly important in the enterprise space where data stored in different data centers, Clouds or application silos can be brought together for a common learning task.

The key design point of {\proj} is fast start-up time for enterprise applications. There is a large library of federated learning algorithms implemented on {\proj} to get users started. Machine learning models tested successfully in a centralized learning application can often be extended to use data in different parties without rewriting the model code, but simply by editing configuration files. Party code is easy to deploy and, depending on the communication layer chosen, might not require opening ports on the party side, which makes deployment of a federated project faster and easier. Using {\proj}, enterprise practitioners can deploy federated learning quickly. In addition, researchers in the field can build on its existing platform to try out new federated learning algorithms and benchmark them against the existing ones in the library.